\documentclass[conference]{IEEEtran}
\IEEEoverridecommandlockouts
\usepackage{cite}
\usepackage{amsmath,amssymb,amsfonts}
\usepackage{algorithmic}
\usepackage{graphicx}
\usepackage{subcaption}
\usepackage{textcomp}
\usepackage{comment}
\usepackage{xcolor}
\usepackage{booktabs}
\usepackage{hyperref}

\def\BibTeX{{\rm B\kern-.05em{\sc i\kern-.025em b}\kern-.08em
    T\kern-.1667em\lower.7ex\hbox{E}\kern-.125emX}}

\begin{document}

\title{Pedestrian Emergency Braking in Ten Weeks
\thanks{Funding from NSF, award no. 1950872}
}

\author{\IEEEauthorblockN{Steven Nguyen}
\IEEEauthorblockA{\textit{Mechanical Engineer} \\
\textit{University of California, Santa Barbara}\\
Santa Barbara, United States \\
stevennguyen@ucsb.edu}
\and
\IEEEauthorblockN{Zillur Rahman}
\IEEEauthorblockA{\textit{Electrical and Computer Engineering} \\
\textit{University of Nevada, Las Vegas}\\
Las Vegas, United States \\
zillur.rahman@unlv.edu}
\and
\IEEEauthorblockN{Brendan Tran Morris}
\IEEEauthorblockA{\textit{Electrical and Computer Engineering} \\
\textit{University of Nevada, Las Vegas}\\
Las Vegas, United States \\
brendan.morris@unlv.edu}

}
\IEEEoverridecommandlockouts
\IEEEpubid{\makebox[\columnwidth]{978-1-6654-7698-0/22/\$31.00~\copyright2022 IEEE \hfill}
\hspace{\columnsep}\makebox[\columnwidth]{ }}

\maketitle
\IEEEpubidadjcol

\begin{abstract}
In the last decade, research in the field of autonomous vehicles has grown immensely, and there is a wealth of information available for researchers to rapidly establish an autonomous vehicle platform for basic maneuvers. In this paper, we design, implement, and test, in ten weeks, a PD approach to longitudinal control for pedestrian emergency braking. We also propose a lateral controller with a similar design for future testing in lane following. Using widely available tools, we demonstrate the safety of the vehicle in pedestrian emergency braking scenarios.
\end{abstract}


\section{Introduction}

While autonomous vehicles (AVs) were once a far-fetched dream only found in science fiction, it has become increasingly clear that they will be integral to the future of our transportation systems. In recent decades, many commercial cars have featured tools to assist drivers, like adaptive cruise control and lane following technology. With all the research in the field, there is a wealth of information available for researchers interested in establishing an AV platform.

The goal of this project was to develop a control algorithm for longitudinal motion of an AV and demonstrate, on a real-world vehicle, the ability to safely stop when approaching a pedestrian. A proportional and derivative gain (PD) controller was designed for control of throttle and braking and was interfaced with the vehicle controller area network (CAN) via the Robot Operating System (ROS). Pedestrians were detected using the YOLOv5 model \cite{glenn_jocher_2022_6222936}. The safety of the proposed longitudinal PD controller was demonstrated by testing the vehicle’s ability to stop at a safe distance from the pedestrian. In addition, it is shown that the proposed controller can stop in a manner that is comfortable for the passengers, by gradually slowing down rather than slamming on the brakes. A similar PD controller was designed for lateral control of the vehicle in lane following, but has not yet been tested in-vehicle.

\section{Related Works}
\label{sec:background}
Challenges in AV control have been tackled using a variety of techniques in the past. One of the most popular techniques for vehicular control is model predictive control (MPC), which excels in nonlinear optimization problems. In \cite{203:MPC_combined_braking_steering_borrellii}, researchers demonstrated that MPC can be used to consolidate the lateral (steering) and longitudinal (braking) controllers in a manner that allows a vehicle to accurately follow a given path despite tortuous conditions of snowy, slippery roads even at high speeds. In addition, MPC was shown in \cite{201:MPC_for_ACC} to be suitable for improving fuel efficiency of AVs. By defining the controller to minimize fuel consumption of the vehicle, researchers increased fuel efficiency by up to 20\%. 

Alternatively, controllers based on fuzzy logic can offer an intuitive approach to controller design that closely mimics human driver behavior \cite{204:Naranjo_fuzzy_brake}. Through the use of fuzzy inference systems, the controller can make inferences on the state and the appropriate control action without necessitating rigorous dynamical models. 

Another approach that has gained popularity is the use of artificial intelligence, through deep neural networks and reinforcement learning \cite{205:lat_long_control_deep_learning}, \cite{206:learning_to_drive_in_day}. Deep learning approaches for AV control often rely on training models to extract features from input images and mapping them to the desired output, which are the desired control actions. Sharma et al. \cite{205:lat_long_control_deep_learning} demonstrated that a deep learning approach, even with limited training data, was capable of navigating tracks in the TORCS simulation environment with minimal lane departure. Alternatively, reinforcement learning is valuable because it converts the challenge of designing the controller into a challenge of training an agent that learns a desirable control law. Kendall et al. \cite{206:learning_to_drive_in_day} presented a reinforcement learning model that can achieve lane-following with just 30 minutes of training. 

Lastly, another common approach to controller design is to use proportional, integral, and derivative (PID) controllers. As one of the simplest to implement control schemes, PID controllers are computationally light and offer a conceptually straightforward design approach. Though PID controllers are not typically as robust as nonlinear approaches, some researchers were able to adopt an adaptive PID controller, where the control gain is tuned in real-time depending on the driving conditions \cite{202:AdaptivePID_for_AV}, to great success. Their approach successfully demonstrated the ability for path tracking and was accurate to within 0.5 meters. Alternatively, PID controllers were designed in a nested loop in \cite{207:Nested_PID_Lateral_Control} for lateral control. In this design, the outer feedback loop is a PID controller to determine the desired yaw rate of the vehicle, and the nested feedback loop is a separate PID controller to stabilize the yaw rate of the vehicle to the reference yaw rate. A similar approach to this is proposed in this study for PD controllers for both longitudinal and lateral control.

In terms of implementation, one consideration that must be made when implementing any control scheme in an AV is the communication between the various sensors and computational units in the vehicle. ROS is a commonly used, open-source software that simplifies this communication by allowing the user to create nodes that publish and read information that can be read and processed by the main CPU. For example, ROS was used in \cite{icab_ros} for implementation in an electric golf cart and in \cite{ros_robot} for an autonomous mobile robot. Alternatively, Guo et. al \cite{MPC_implement} used User Datagram Protocol (UDP) to handle the communication between processing units when implementing MPC in an AV. UDP is used in lieu of ROS by Jones et. al \cite{senior_implement} for its favorable speed in an implementation that was similar in scope to this work. However, their work is more focused on simulations and neural network training whereas this work will focus on the implementation of a closed loop control scheme. In this work, ROS is chosen due to its widespread use and support for AVs for rapid development.

\section{Control Schemes} \label{sec:control}
The control basics consists of a longitudinal controller to regulate acceleration and braking and a lateral controller to handle steering of the vehicle.

\subsection{Longitudinal Control}
For the challenge of braking in the presence of pedestrians, it is of utmost importance that the longitudinal controller ensures the vehicle will stop at a safe distance. To assist in this, the proposed control algorithm subtracts a predetermined distance from the pedestrian's position to use as the stopping position. This distance can be changed depending on the driver's comfort. Position and velocity feedback are both used to prevent overshoot on the desired stopping position and to ensure the vehicle gradually stops in a manner comfortable for the passengers.

\begin{figure}[t]
\centerline{\includegraphics[width=0.95\linewidth]{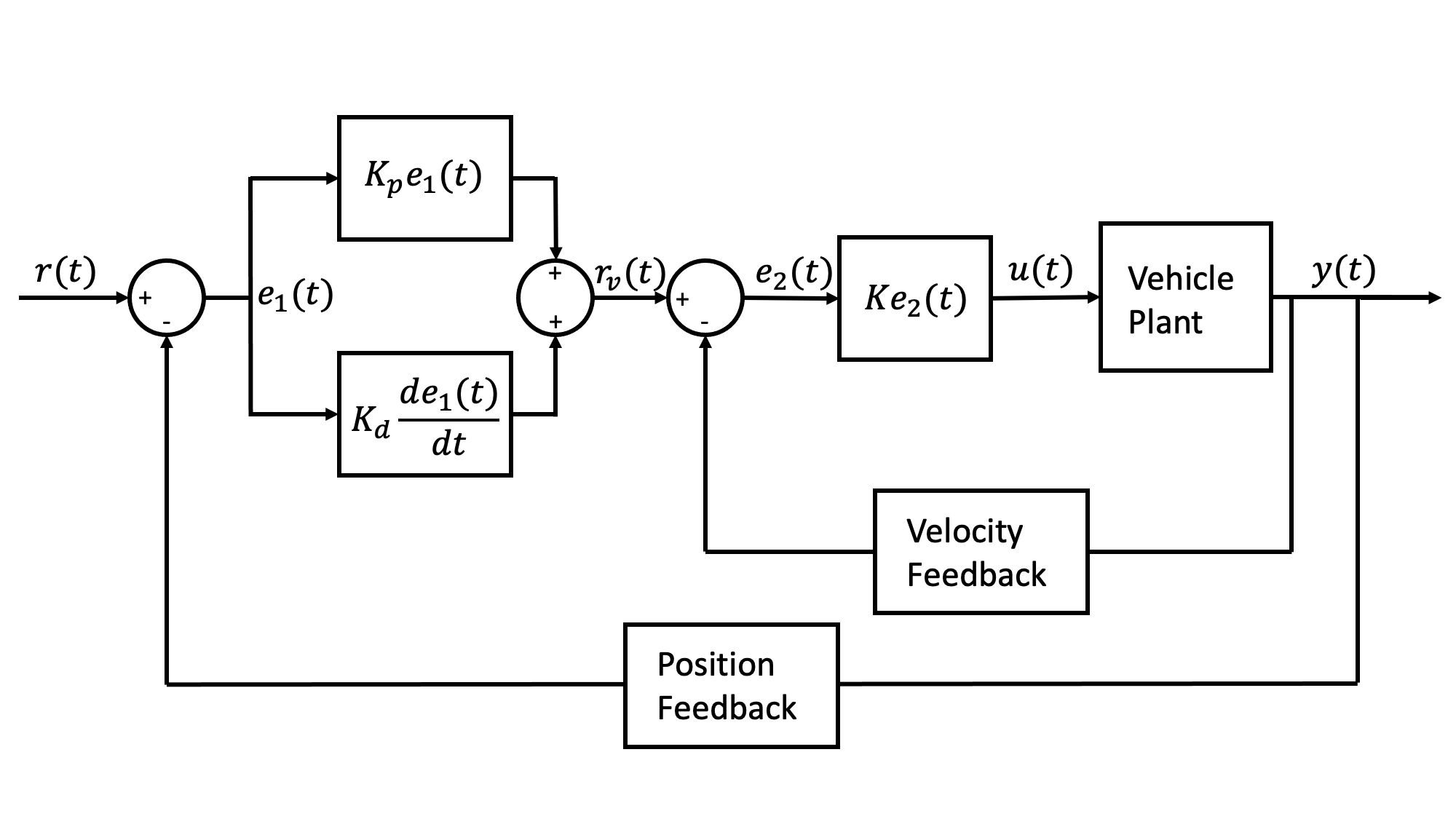}}
\caption{Block diagram for longitudinal control algorithm with both position and velocity feedback.}
\label{fig:Nested_diagram}
\end{figure}

The PD controller design is shown in the block diagram in Fig. \ref{fig:Nested_diagram}. The outer feedback loop uses position feedback and proportional and derivative control to generate a reference velocity, $r_v(t)$, 
\begin{equation}
    r_v(t) = K_p e_1(t) + K_d \frac{d}{dt}e_1(t).
    \label{eq:r_v}
\end{equation}
This reference velocity is then used in the inner feedback loop, which is designed to stabilize the velocity of the vehicle and gradually bring it to a stop using a proportional gain, $K$. The inner loop generates the braking force, $u(t)$, 
\begin{equation}
    u(t) = K e_2(t),
    \label{eq:u}
\end{equation}
which is converted into a braking command for the vehicle.

The combination of these feedback loops ensures that, as the vehicle approaches the referenced stopping position, $r(t)$, the reference velocity similarly approaches 0, and the vehicle comes to a stop. This approach is similar to that used by Marino et al. in \cite{207:Nested_PID_Lateral_Control}.

For the stability analysis, a simple model of the vehicle longitudinal dynamics was used for the plant as shown below:
\begin{equation}
    \frac{d}{dt}\left[ \begin{array}{c} x \\ \dot{x} \end{array} \right] = \left[\begin{array}{cc}0 & 1 \\ 0 & 0 \end{array}\right] \left[ \begin{array}{c} x \\ \dot{x} \end{array} \right] + \left[ \begin{array}{c}0 \\ \frac{1}{m} \end{array}  \right] u.
    \label{eq:plant}
\end{equation}
It accounts for the inertia from the vehicle mass, $m$, and treats the braking force as the input. The car's position is the variable $x$. Drag was excluded in stability analysis because of its nonlinear dependence on the square of velocity and its minimal impact at the velocities during testing (under 25 miles per hour). In practice, drag serves to increase the braking force, and increases the safety margin. Drag effects are included in the simulations in Section \ref{sec:results}.

The complementary sensitivity function from the reference position to the vehicle position is given as: 
\begin{equation}
    G_{\textrm{Position,}R}(s) = \frac{K \left(K_p + K_d s \right)}{ms^2 + K K_d s + K + K K_p},
    \label{eq:tf_long}
\end{equation}
where $K$, $K_p$, and $K_d$ represent the controller gains as depicted in the block diagram, and $m$ is the mass of the vehicle. By definition, all the coefficients of the denominator polynomial are positive, so the system is stable by the Routh-Hurwitz stability criterion. 

While a PD controller with position feedback would have been sufficient to ensure stability of the closed-loop system, velocity feedback was incorporated in the proposed manner to address the requirement of comfort for passengers. The sensitivity function for the velocity feedback loop is given as:
\begin{equation}
    G_{E_2(t),R_v}(s) = \frac{ms}{ms + K}.
    \label{eq:velocity_sens}
\end{equation}
By the final value theorem, the error of the vehicle velocity with respect to the reference velocity, $e_2(t)$, is bounded in ramp response. This is significant because it means that for any given pedestrian distance and initial velocity, the acceleration of the car can be tuned depending on the reference velocity. With the proposed controller, this means that the gains $K_p$ and $K_d$ can be tuned to increase or decrease the maximum acceleration of the vehicle. 

\begin{figure}[t]
\centerline{\includegraphics[width=0.95\linewidth]{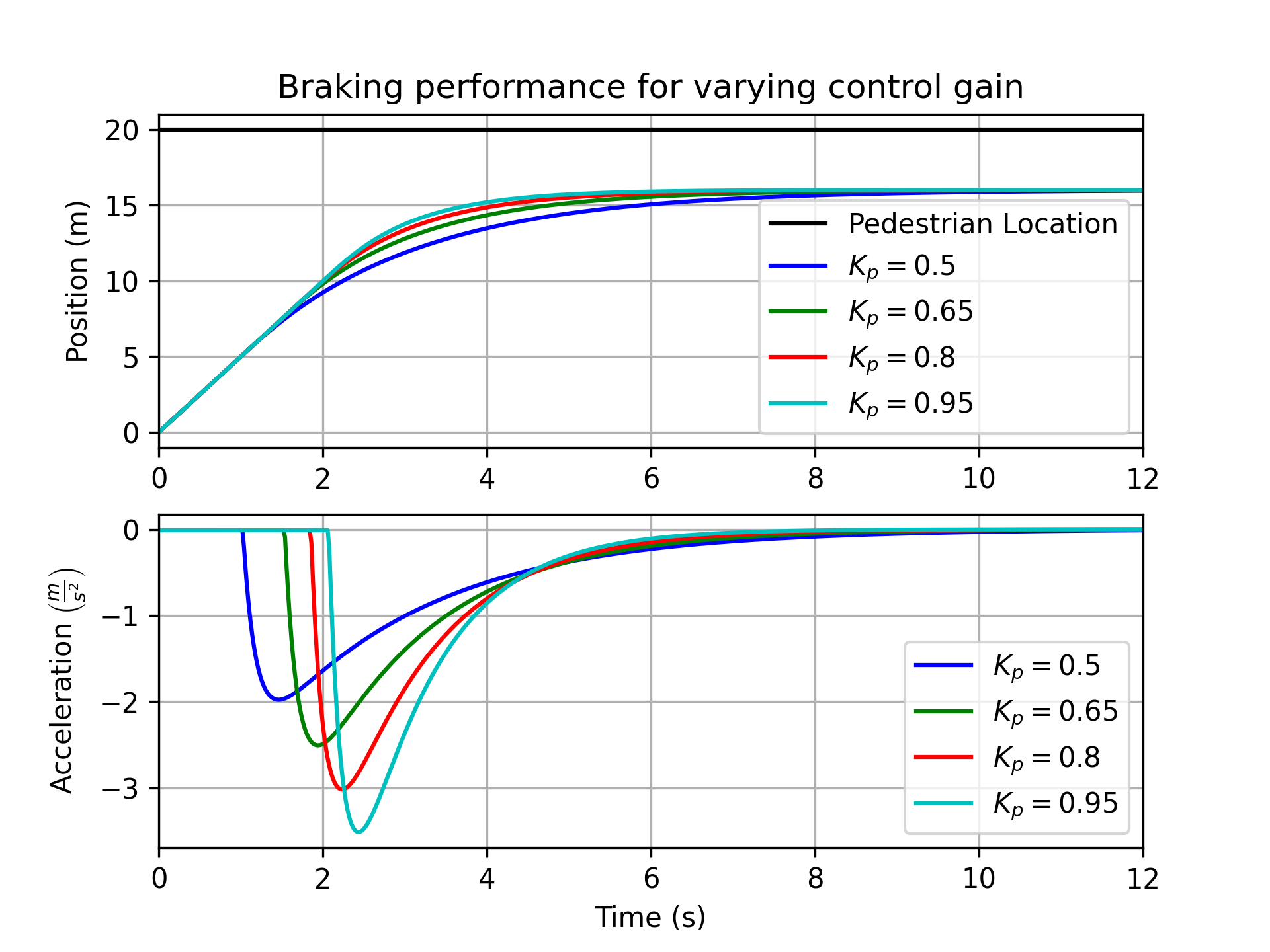}}
\caption{This design offers control over the acceleration that the passenger experiences by tuning $K_p$.}
\label{fig:changing_k_p}
\end{figure}

In Fig. \ref{fig:changing_k_p}, the effect of varying $K_p$ is shown. Lowering $K_p$ results in a lower acceleration because the vehicle will begin braking earlier and will more gradually approach the stopping point. This extra control over the acceleration of the vehicle does not sacrifice any safety, as all of the trajectories converge to the same final position. The experimental behavior of this control system and the simulated responses are shown in Section \ref{sec:results}.

For the experiments conducted in Section \ref{sec:results}, gain values of $K_p = 0.8$, $K_d = 0.1$, and $K = 10,000$ were used. The large value for $K$ is because the force $u(t)$ accounts for the vehicle mass of approximately 1,725 kilograms.

\subsection{Lateral Control}
A lateral control algorithm is also proposed with a similar design to the longitudinal controller. Whereas longitudinal control governs how a vehicle will brake or accelerate, lateral control governs the steering and yaw rate of the vehicle.

\begin{figure}[t]
\centerline{\includegraphics[width=0.95\linewidth]{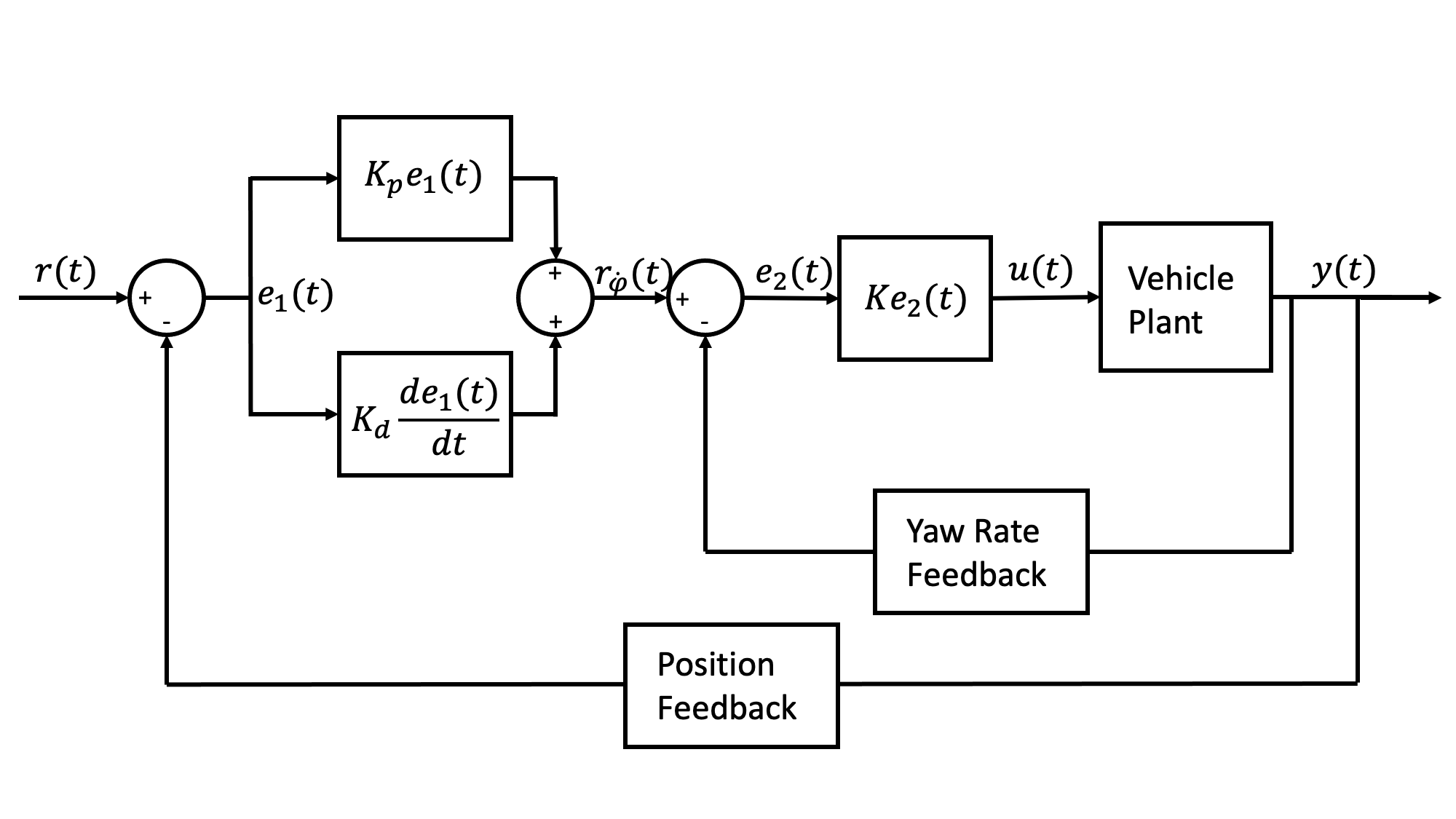}}
\caption{Proposed control algorithm uses both position and velocity feedback.}
\label{fig:lateral_diagram}
\end{figure}

The block diagram for the lateral controller is shown in Fig. \ref{fig:lateral_diagram}. It is identical to the longitudinal controller in most respects, but different because the inner loop is used for feedback control of the yaw rate of the vehicle rather than the velocity. Instead of a reference velocity, $r_v(t)$, there is a reference yaw rate, $r_{\dot{\psi}}$.

To model the vehicle yaw rate dynamics, the linearized model presented by Antonov et al. \cite{13:Best_lateral_dynamics} was used. The state space model is
\begin{align}
    \frac{d}{dt} \left[  \begin{array}{c} v_y \\ \dot{\psi} \end{array} \right] = & \left[ \begin{array}{cc} -\frac{c_r + c_f}{mv_x} & \frac{c_r l_r - c_f l_f}{mv_x} -v_x \\ \frac{c_r l_r - c_f l_f}{I_z v_x} & -\frac{c_r l_r^2 + c_f l_f^2}{I_z v_x} \end{array} \right] \left[  \begin{array}{c} v_y \\ \dot{\psi} \end{array} \right] + \nonumber \\
    &\left[\begin{array}{c} \frac{c_f}{m} \\ \frac{c_f l_f}{I_z}   \end{array} \right]\delta_f + \left[\begin{array}{c} 0 \\ \frac{1}{I_z}  \end{array}   \right] M_z.
    \label{eq:lateral_ss}
\end{align}
The variables $c_f$ and $c_r$ represent the cornering stiffnesses of the front and rear axles, $m$ is the mass, $v_x$ is the longitudinal velocity of the vehicle, $v_y$ is the lateral velocity of the vehicle, and $l_r$ and $l_f$ are the distances of the rear and front axles from the vehicle center of mass \cite{13:Best_lateral_dynamics}. $M_z$ is the yaw torque, but is taken to be 0 in our model. Stability analysis was not performed on the lateral control algorithm because the dynamical model of the vehicle's position as a function of the yaw rate was a nonlinear equation.

\begin{figure}[t]
\centerline{\includegraphics[width=0.95\linewidth, height=6cm]{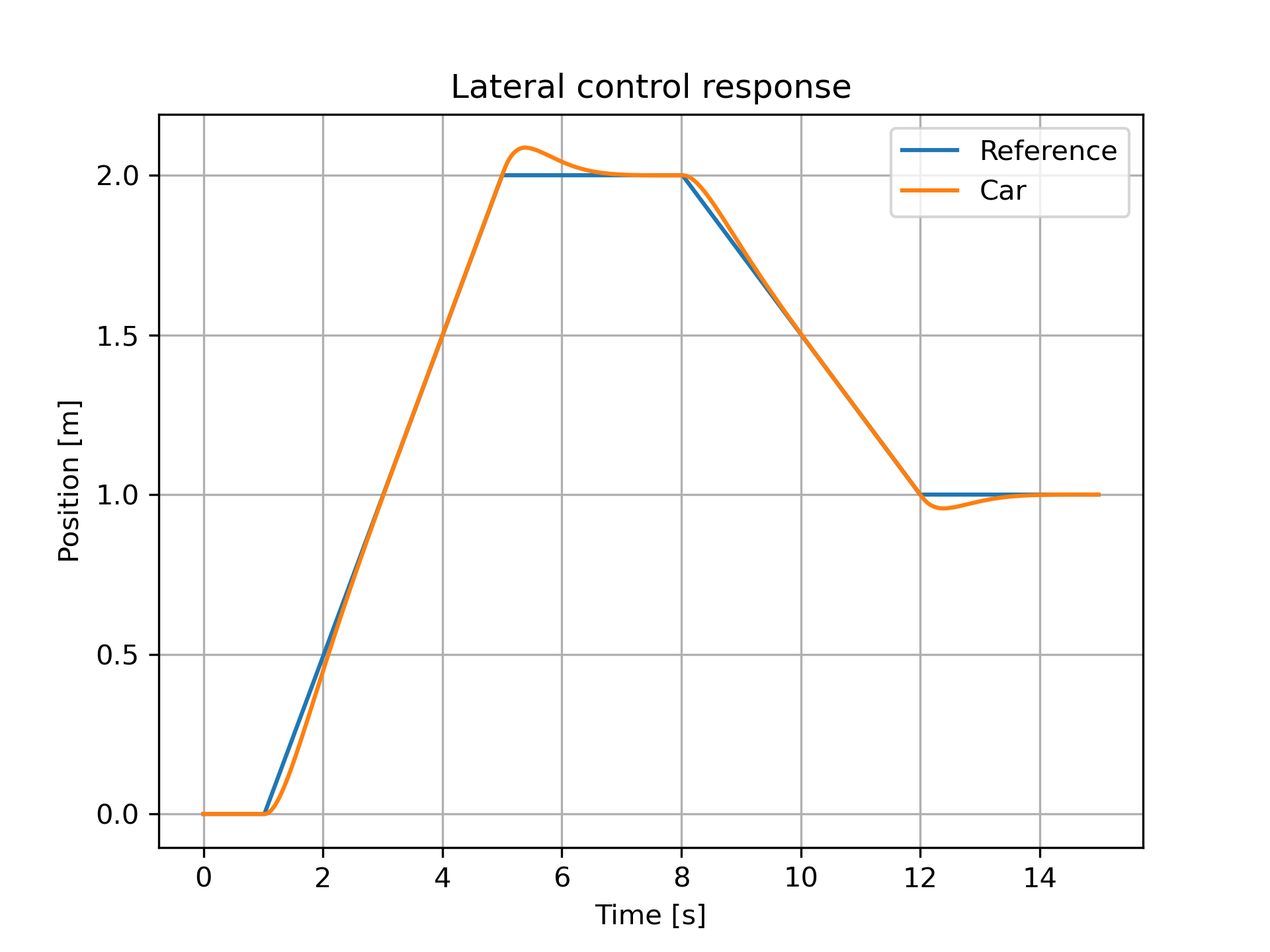}}
\caption{Response of lateral control algorithm shows that it is stable in step response.}
\label{fig:lateral_response}
\end{figure}

While the lateral control algorithm was ultimately not tested on the vehicle, the simulated response of the system to a varying reference position is shown in Fig. \ref{fig:lateral_response}. The response of the closed-loop system is stable in step response. While there is some overshoot, it is negligible compared to the width of an average road lane.

\section{Car Setup and Implementation} \label{sec:setup}

The pedestrian brake controller was implemented on a vehicle for real-world experimentation.  Table \ref{tab:resources} consolidates information on the vehicle setup.

\subsection{Drive-by-Wire System}
The vehicle platform is a 2017 Lincoln MKZ Hybrid, fitted with a drive-by-wire system provided by Dataspeed \cite{dataspeed}. ROS is used to communicate with the vehicle controller area network (CAN). The CAN system signals are decoded using Dataspeed's algorithms, so information from the vehicle sensors can be read and commands can be sent to the vehicle from a computer. It is through the CAN system that we are capable of applying feedback control to drive the vehicle. The drive-by-wire system included a Logitech F310 joystick controller that can be used to electronically control the vehicle. 

For safety purposes, one researcher was always present in the driver's seat to take control of the car if necessary. Manual inputs to the steering wheel or brakes were hard-wired to disable the drive-by-wire system immediately.

\subsection{Vehicle camera setup}
\begin{figure*}[t]
\centering
\begin{subfigure}[b]{0.45\textwidth}
\includegraphics[width=1.0\linewidth]{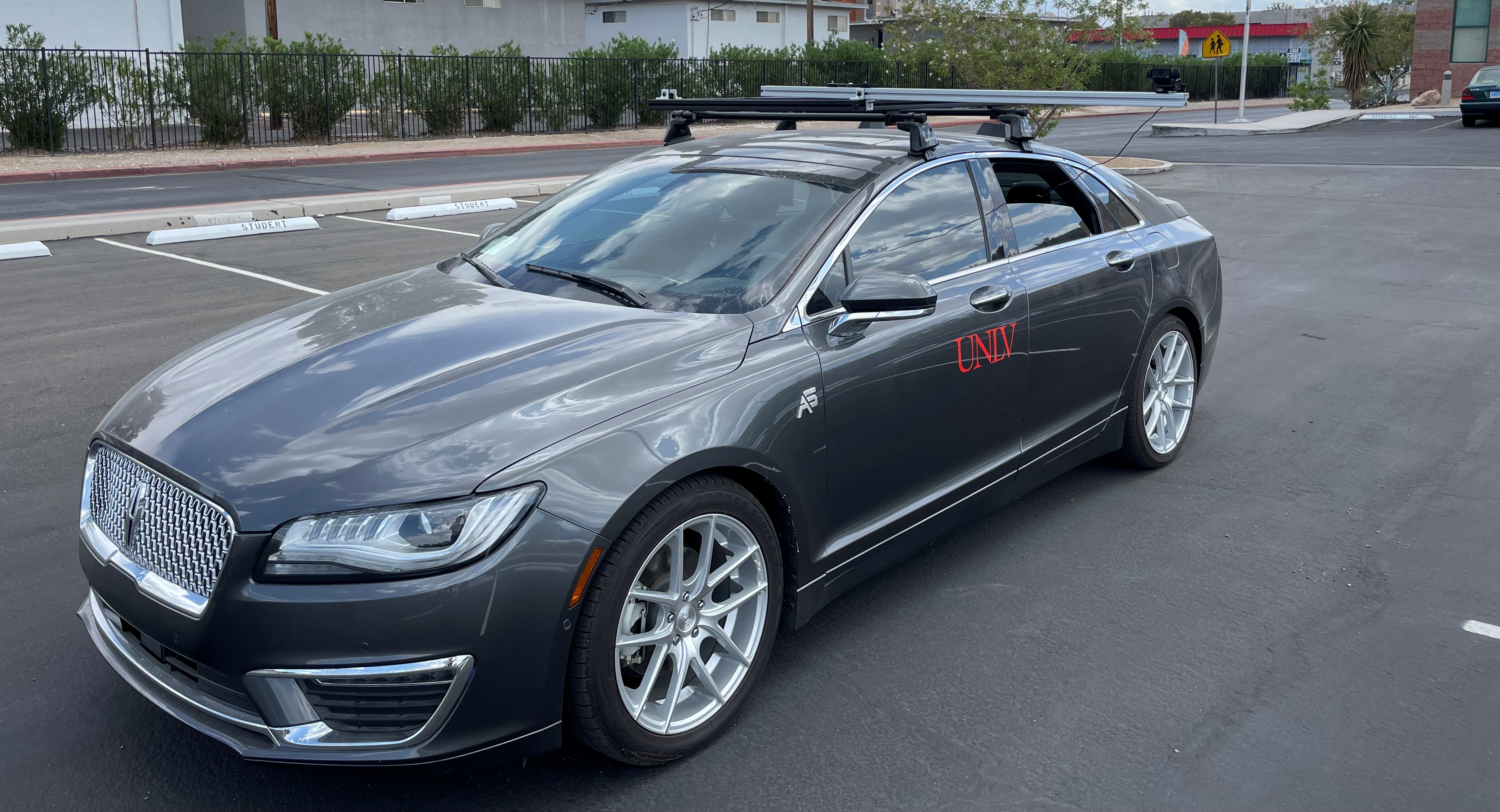}
\caption{}
\label{fig:cam1}
\end{subfigure}
\begin{subfigure}[b]{0.45\textwidth}
\includegraphics[width=1.0\linewidth]{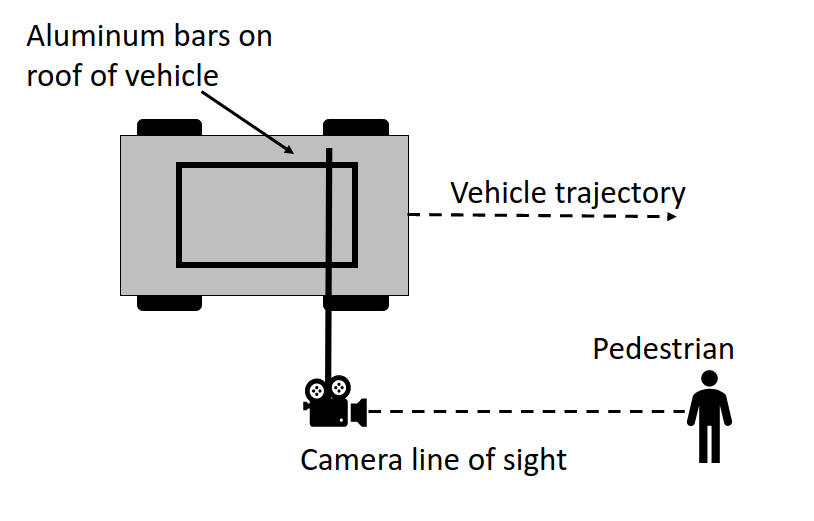}
\caption{}
\label{fig:cam2}
\end{subfigure}
\caption{(\subref{fig:cam1}) Picture of vehicle setup with camera mounted. (\subref{fig:cam2}) Diagram of test with pedestrian offset from vehicle trajectory. }
\label{fig:car_setup}
\end{figure*}

To mount the camera on the vehicle, a rack of extruded aluminum bars was attached to the roof of the vehicle. The bar that the camera was attached to was mounted such that it could be slid to the side of the vehicle in order to hold the camera at an offset distance from the side of the vehicle. This setup and the positioning of the pedestrian relative to the vehicle can be seen in Fig. \ref{fig:car_setup}. The purpose of this setup was to allow for safer testing with pedestrians. With the camera protruding from the side of the vehicle, the pedestrian could stand clear of the vehicle's trajectory while they were detected as if they were in front of the vehicle.

\subsection{Pedestrian Detection Model}
\begin{figure}[t]
\centerline{\includegraphics[width=0.95\linewidth]{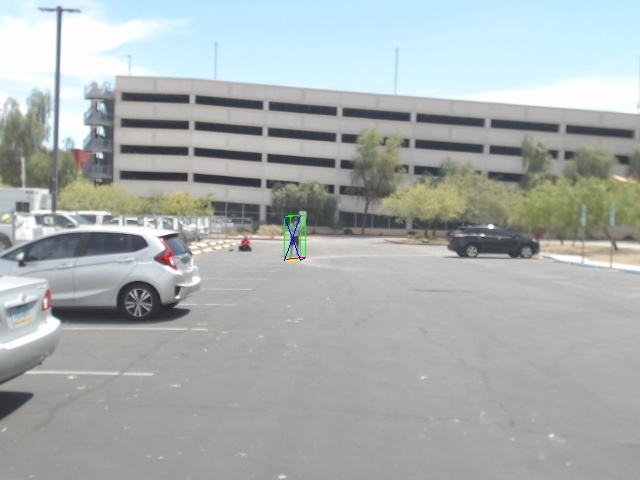}}
\caption{Sample pedestrian detection at \~20m.}
\label{fig:pedestrian}
\end{figure}

For the pedestrian detection and distance estimation from a single RGB camera feed, a monocular 3D detection model \cite{3d_det} was used. In this model, a pre-trained VGG network \cite{simonyan2014very} without dense layers was used to regress 3D object properties. Then, these estimates were combined with the geometric constraints provided by a YOLOv5 \cite{glenn_jocher_2022_6222936} 2D object detector. This model provides the distance of the detected object which was then sent to the ROS longitudinal control node along with current car velocity to generate the appropriate brake commands. Fig. \ref{fig:pedestrian} shows a sample detection where the person is approximately 20 meters away from the camera.

\subsection{Lane Detection Model}
\begin{figure}[t]
    \centering
    \includegraphics[width=0.95\linewidth]{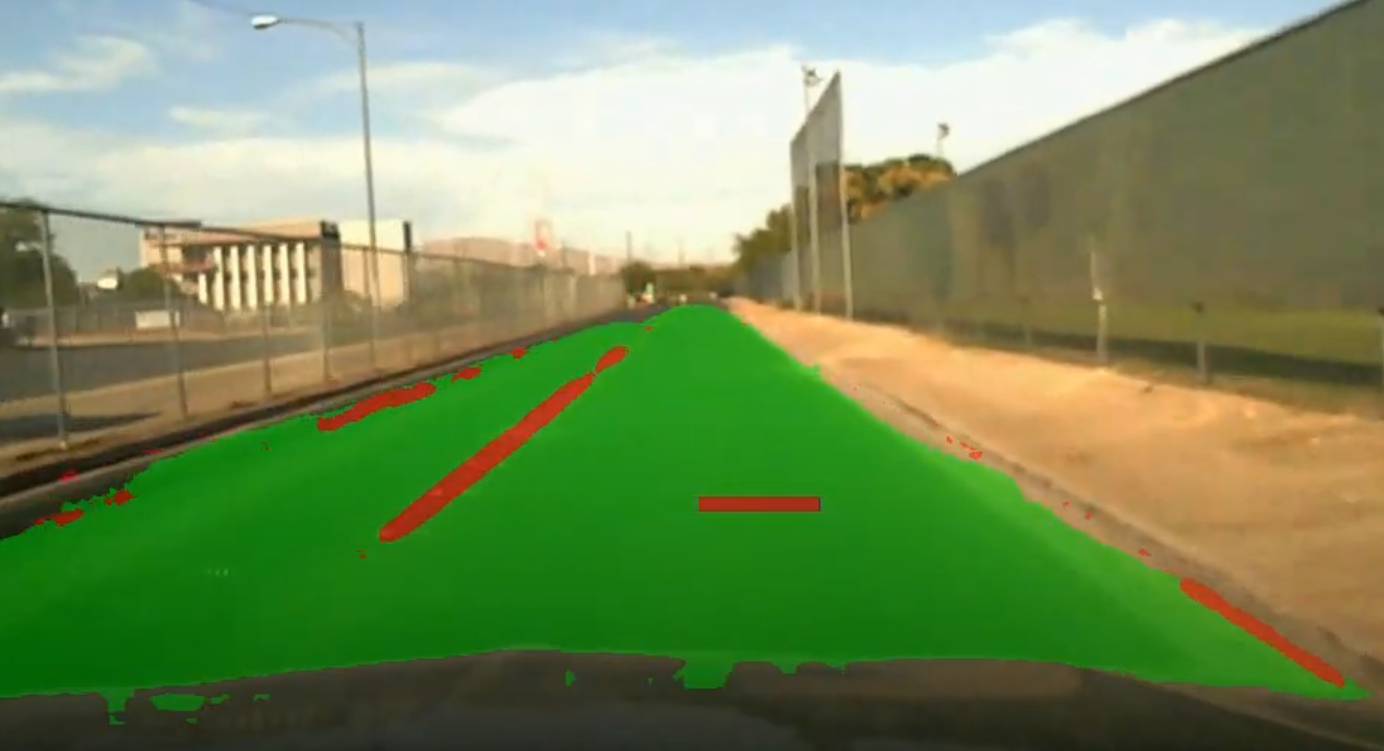}
    \caption{YOLOP lane detection on UNLV campus road}
    \label{fig:kalman_filter}
\end{figure}

The YOLOP model \cite{YOLOP} was investigated for its lane detection capabilities. YOLOP is a driving perception model that is capable of traffic object detection, drivable space segmentation, and lane detection. By default, the YOLOP demo program will output the drivable space and lane markings. In order to track the center of the lane, a Kalman filter was developed, and its output shown in Fig. \ref{fig:kalman_filter} as the horizontal red line. The green color and red lane lines in the image were outputs of the default demo program. The center-of-lane detection works simply by finding the left and right bounds of the road and taking averages to find the center of the road and the center of the lane. While this was not used for a test, it may be useful for future work with the lateral control algorithm, where the vehicle is tasked with lane following.

\subsection{ROS Setup}
For this experiment, we used ROS1 in an Ubuntu 20.04 computer equipped with an Nvidia RTX 2070 GPU. For the programming languages, both C++ and Python were used. One ROS node received the raw RGB images and published them as ROS ``Image'' messages which were then received by the pedestrian detection node. The object distance published by that node was sent to the longitudinal control node. The longitudinal controller as described in section \ref{sec:control} generates the brake level based on current velocity and distance. Generally, the greater the velocity, the earlier it generates brake command to avoid possible collision. The brake message was then converted to ROS ``CAN'' messages and sent to the Dataspeed CAN system. 


\begin{table}[t]
\caption{AV Platform Resources}
\label{tab:resources}
\centering
\begin{tabular}{rl}
\toprule
\textbf{Resource} & \textbf {Description} \\
\midrule
Vehicle & 2017 Lincoln MKZ Hybrid \\
Drive-by-Wire & \href{https://bitbucket.org/DataspeedInc/dbw_mkz_ros/}{Dataspeed}\\
Camera & Logitech C920 Webcam \\
Joystick Controller & Logitech F310 \\
Computer Specs & Intel Xeon E5-2603 and Nvidia RTX 2070 GPU \\
\href{https://www.ros.org/}{ROS} & Open source robotics library\\
Dataspeed ROS Driver & \href{https://bitbucket.org/DataspeedInc/dbw\_mkz\_ros/}{dbw\_mkz\_ros}\\
Pedestrian Detection & \href{https://github.com/ultralytics/yolov5}{YOLOv5} and \href{https://github.com/skhadem/3D-BoundingBox}{3D Object Detection}\\

\bottomrule
\end{tabular}
\end{table}

\section{Results and Discussion} \label{sec:results}
Experiments were designed to verify the operation of subcomponents before testing with a pedestrian on the road. First, the ability of the controller to stop the car was verified by generating a synthetic pedestrian signal at a specified distance and verifying that braking resulted in stopping at the safety distance. After confirming the controller's ability to reliably stop, the pedestrian detection program was characterized, and finally a full system in-car pedestrian emergency brake test was performed.

\subsection{Pedestrian Detection}
\begin{figure}
    \centering
    \includegraphics[width=0.95\linewidth]{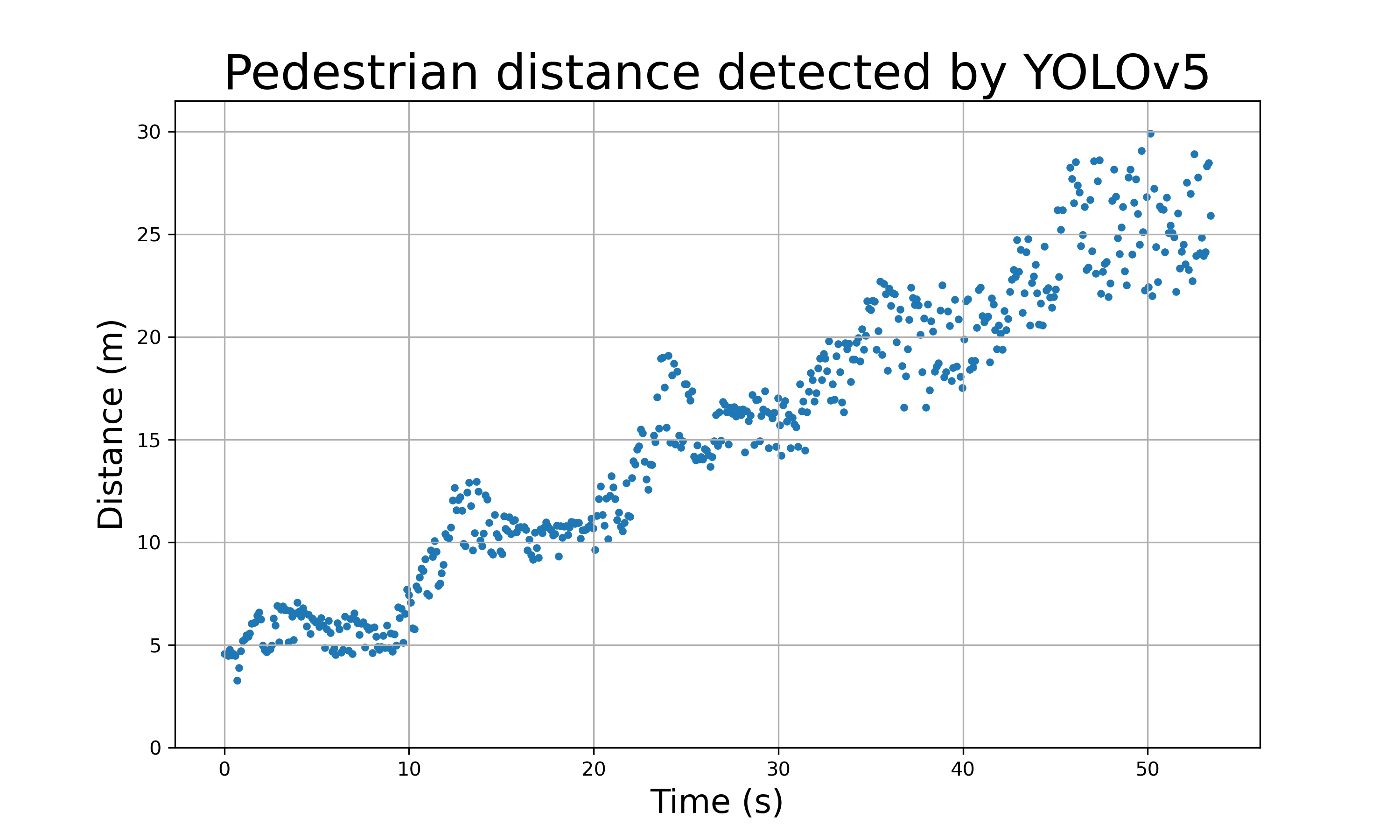}
    \caption{Raw pedestrian detection with stationary vehicle and moving pedestrian.}
    \label{fig:detection_range}
\end{figure}

The quality of distance measurement was evaluated with the vehicle stationary and having a moving pedestrian to test the accuracy of the detection at various ranges. The detection results are shown in Fig. \ref{fig:detection_range}. The pedestrian stood at 5, 10, 15, 20, and 25 meters from the vehicle camera for approximately 5 seconds each during this test.

As shown in the plot, the precision of the raw detection decreased as the distance from the camera increased, though noise was clearly evident throughout the entire test range.  The noise is unsurprising since the pedestrian distance is estimated using a monocular 3D algorithm.  However, this noise was problematic for the braking tests and is discussed below.

\subsection{Braking Performance with Pedestrian}
\begin{figure*}[h]
    \centering
    \begin{subfigure}[b]{0.3\textwidth}
        \includegraphics[width=\textwidth]{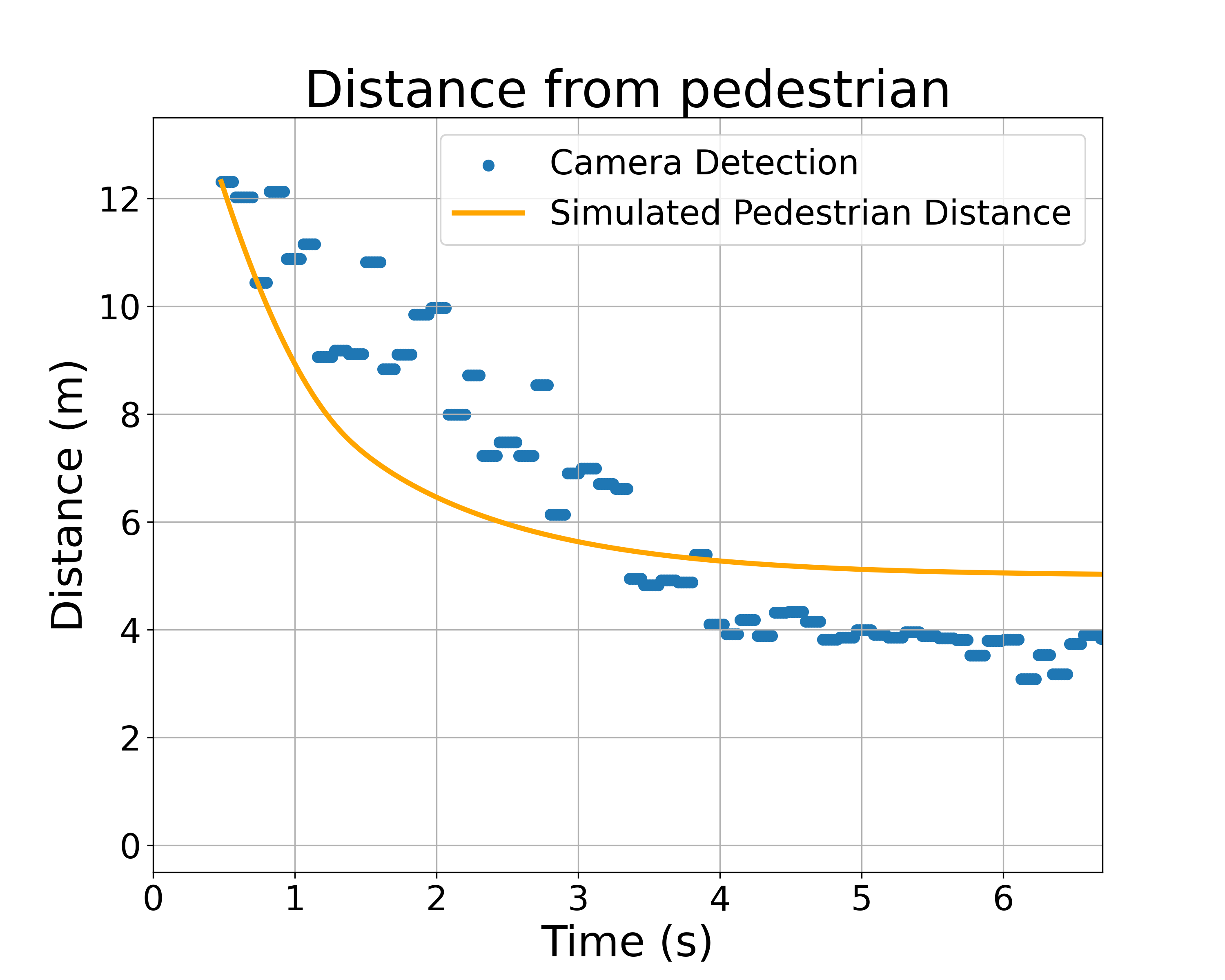}
        \caption{}
        \label{fig:dist_comp}
    \end{subfigure}
    \begin{subfigure}[b]{0.3\textwidth}
        \includegraphics[width=\textwidth]{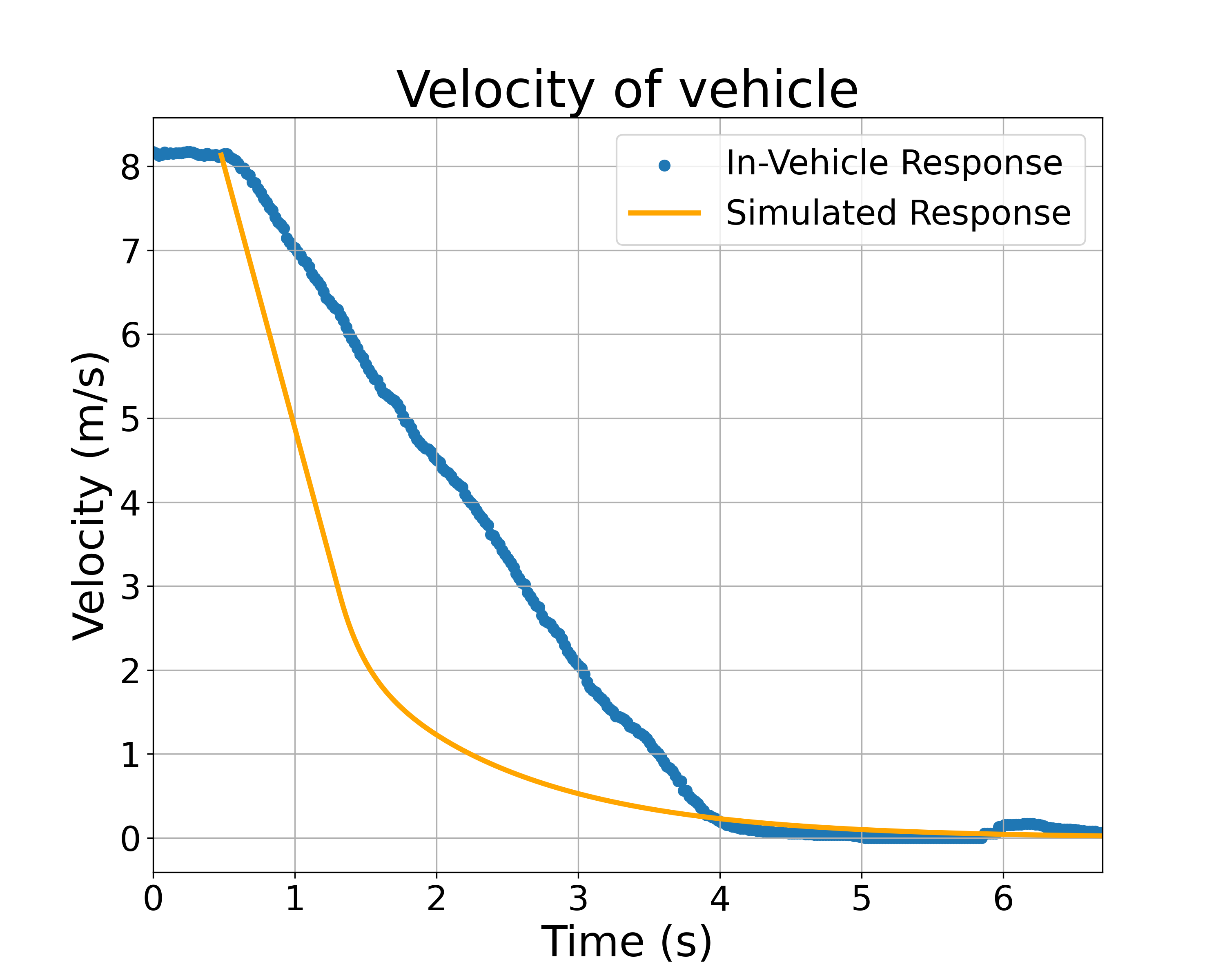}
        \caption{}
        \label{fig:vel_comp}
    \end{subfigure}
    \begin{subfigure}[b]{0.3\textwidth}
        \includegraphics[width=\textwidth]{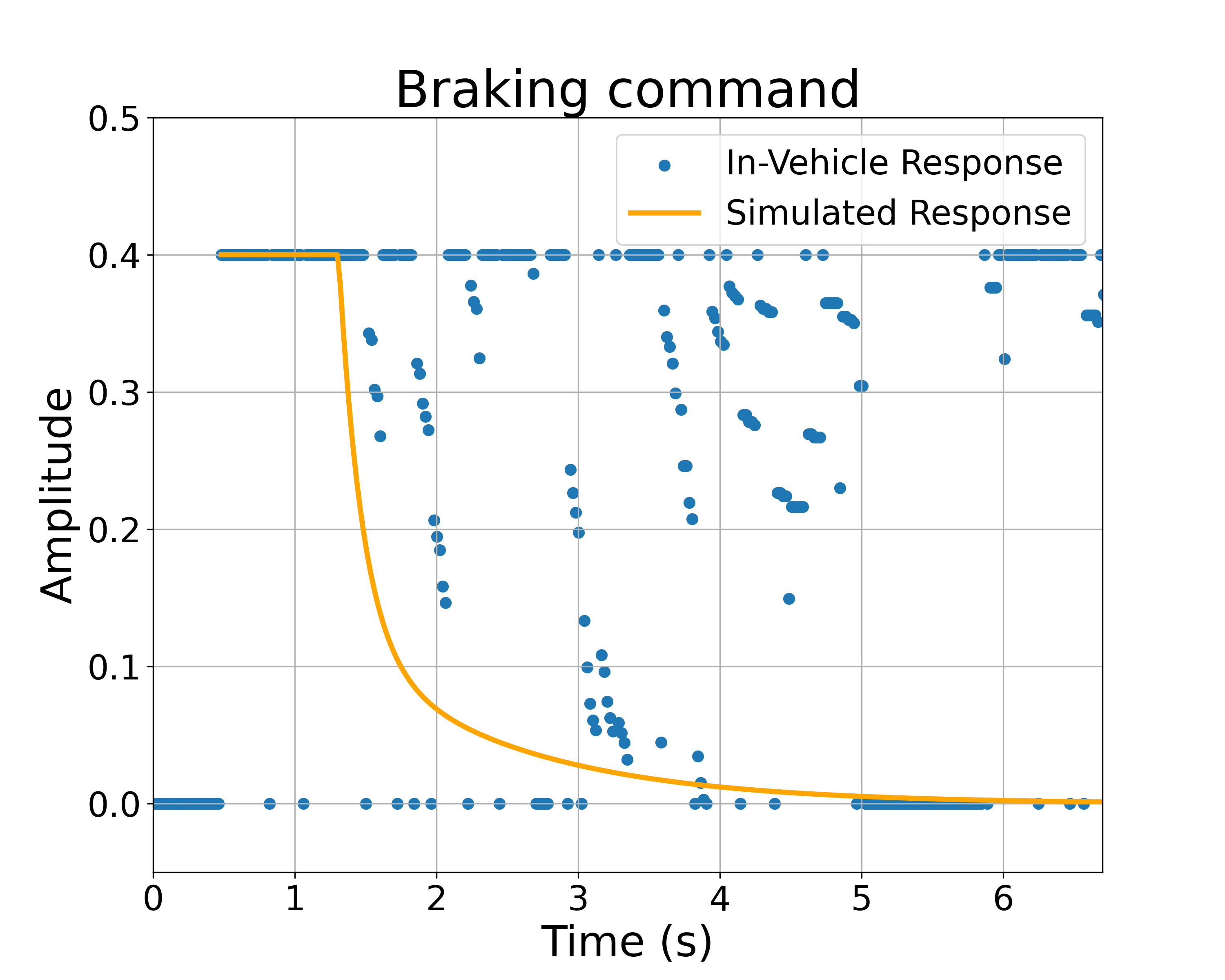}
        \caption{}
        \label{fig:brake_comp}
    \end{subfigure}
    \caption{Comparison between in-vehicle and simulated response for braking test. (\subref{fig:dist_comp}) Distance comparison, (\subref{fig:vel_comp}) velocity comparison, (\subref{fig:brake_comp}) brake command comparison}
    \label{fig:3plot_comparison}
\end{figure*}

The control algorithm was tested in the vehicle setup described in Section \ref{sec:setup}. For this test, the vehicle was driven via the Logitech F310 controller to accelerate up to a speed of approximately 8.13 meters per second, or approximately 18 miles per hour, and the brake commands were generated independently by the longitudinal control algorithm. The results of this test were compared with simulated results shown in Fig. \ref{fig:3plot_comparison}. In both the simulated and in-vehicle results, the vehicle was specified to stop 5 meters from the pedestrian.

As evidenced by the figure, the simulated case responds more quickly to the detected pedestrian and stops approximately one meter earlier than the in-vehicle test, at exactly five meters. In addition, the simulated response is smoother than the in-vehicle response. These are likely results of the noisy pedestrian signal. The noise in the pedestrian distance signal during the in-vehicle test can be seen in Fig. \ref{fig:dist_comp}. Without a smooth detection of the pedestrian, the control algorithm will be unable to properly generate the appropriate braking command, which may be why the in-vehicle test stopped closer to the pedestrian than the simulated test.

In addition, while the in-vehicle velocity appears smooth and linear, the braking was not smooth and rapidly engaged and disengaged. The brake commands are compared in Fig. \ref{fig:brake_comp}, where the jittery, in-vehicle braking can be observed. This is most likely another problem introduced by the noisy detection. The comparison of the brake commands also shows that when the vehicle first starts braking, both the simulated and actual vehicle use a brake command of 0.4, but their accelerations shown in Fig. \ref{fig:vel_comp} do not match. When converting the brake force $u(t)$ into the command signal for the vehicle, the control algorithm uses information from Dataspeed that characterized the braking and acceleration of the vehicle. However, the discrepancy between the simulated and actual accelerations may warrant additional testing to verify the brake torque, or a more nuanced model of the vehicle braking.

Despite the issues introduced by the noisy input, the vehicle was able to safely stop with ample distance from the pedestrian. In other tests, it was found that this stopping distance could be tuned to have the vehicle stop farther or closer to the pedestrian as well.

\section{Future Works} \label{sec:future}
For future work, the Lux LiDAR and Delphi ESR RADAR in the car will also be used in tandem with the RGB camera to more accurately estimate the pedestrian position. A tracking method will also be applied to more smoothly track the pedestrian distance. With a smoother and more accurate pedestrian detection, it will be easier to more rigorously test properties of the controller, like its reliability at various speeds and comfort in braking.

Looking beyond the longitudinal controller, future research will test the proposed lateral controller for its ability in lane following and implement it alongside the longitudinal controller for full autonomous control.

\section{Conclusion} \label{sec:conclusion}
By building off of the wealth of research in the field of AVs, a controller for pedestrian emergency braking was designed and tested in a matter of weeks. We began designing the controller by reviewing the various approaches to controller design and selected a PID approach for its simplicity and ease of implementation. After simulating the results and testing various components, we were able to implement it in a vehicle with a drive-by-wire system and demonstrated safe braking. In addition, a lateral controller was designed that will be useful for ensuring the vehicle can navigate lanes properly and increase the degree of autonomy. 

\bibliographystyle{IEEEtran}
\bibliography{main}

\end{document}